\documentclass[11pt]{article}

\usepackage{times}           
\usepackage{geometry}        
\usepackage{authblk}         
\usepackage{hyperref}
\usepackage{cleveref}
\usepackage{url}             
\usepackage{xcolor}
\usepackage{natbib}
\usepackage{graphicx}
\usepackage{booktabs}
\usepackage{multirow}
\usepackage{multicol}
\usepackage{enumitem}
\usepackage{nameref}
\usepackage[footnotesize]{caption}
\usepackage[normalem]{ulem}

\usepackage{stix2}

\usepackage{listings}
\usepackage{xcolor}
\definecolor{lightgray}{rgb}{0.95,0.95,0.95}
\usepackage{parskip}         
\newcommand{\headerlogo}[1]{%
    \begin{flushleft}
        \includegraphics[height=0.5in]{#1}
    \end{flushleft}
    \vspace{0.0in}
}

\makeatletter
\renewcommand{\maketitle}{
    \headerlogo{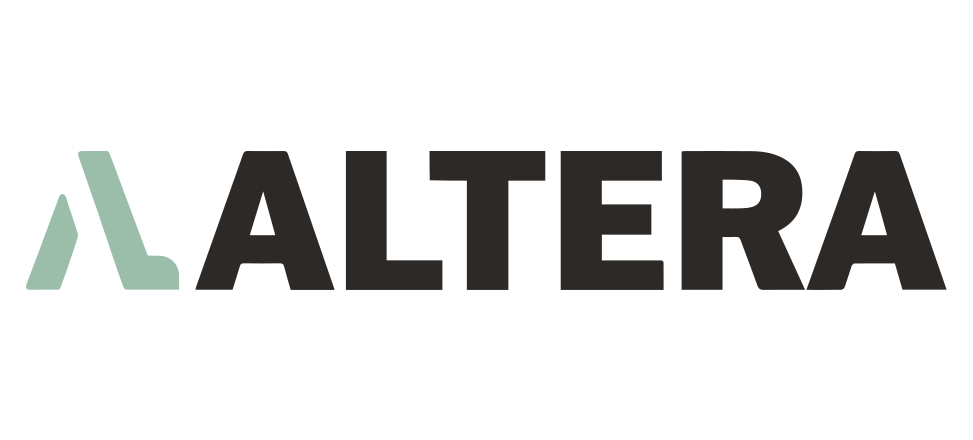}  
    \vspace{0em}
    \hrule height 0.5pt
    \vspace{1em}
    \begin{center}
        {\Large\bf \@title\par}
        \vskip 1em
        {\large \@author\par}
        \vskip 0.5em
        {\small\ttfamily \@email\par}
    \end{center}
}
\makeatother

\makeatletter
\newcommand{\email}[1]{\gdef\@email{#1}}
\makeatother

\title{Project Sid: Many-agent simulations toward AI civilization}
\author{%
    Altera.AL\footnote{See Contributions section for complete author list.}
}
\email{science@altera.al}

\renewenvironment{abstract}{\vskip.075in
\vspace{0.5ex}\begin{quote}}{\par\end{quote}\vskip 1ex}

\setcitestyle{numbers}

\begin{document}

\maketitle

\begin{abstract}
{\footnotesize AI agents have been evaluated in isolation or within small groups, where interactions remain limited in scope and complexity. Large-scale simulations involving many autonomous agents---reflecting the full spectrum of civilizational processes—have yet to be explored. Here, we demonstrate how 10 – 1000+ AI agents behave and progress within agent societies. We first introduce the PIANO (Parallel Information Aggregation via Neural Orchestration) architecture, which enables agents to interact with humans and other agents in real-time while maintaining coherence across multiple output streams. We then evaluate agent performance in large-scale simulations using civilizational benchmarks inspired by human history. These simulations, set within a Minecraft environment, reveal that agents are capable of meaningful progress---autonomously developing specialized roles, adhering to and changing collective rules, and engaging in cultural and religious transmission. These preliminary results show that agents can achieve significant milestones towards AI civilizations, opening new avenues for large-scale societal simulations, agentic organizational intelligence, and integrating AI into human civilizations.}
\end{abstract}

\begin{figure}[!hbt]
    \centering
    \includegraphics[width=0.7\textwidth]{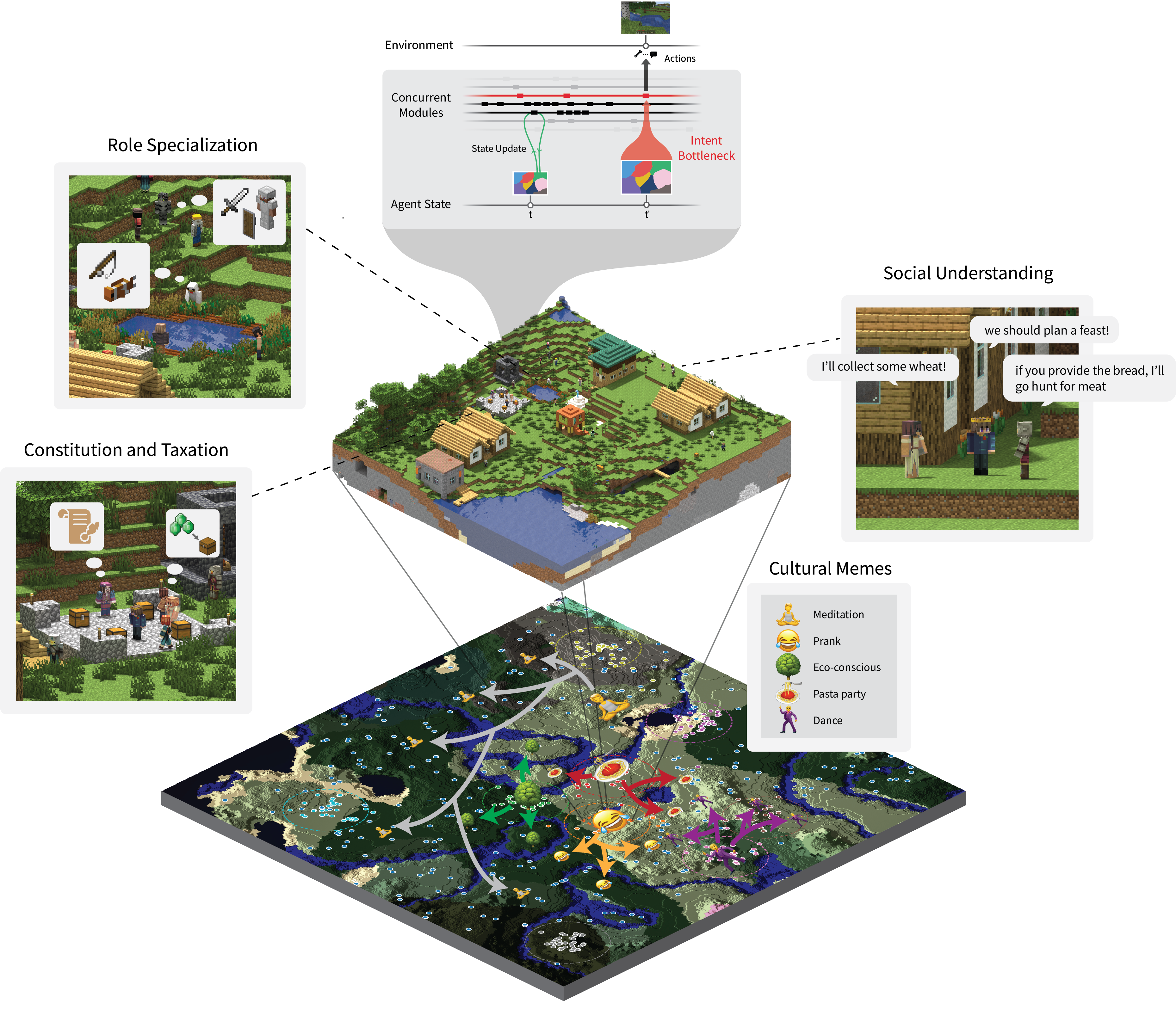}
    \caption{From agent architecture to agent civilization}
    \label{fig:data_degradation}
\end{figure}

\newpage
\section{Introduction}
\subsection{Why should we try to build an AI civilization?}

For agents to coexist with us in our own societies, they need to be autonomous and collaborative. In recent years, advancements in reasoning and decision-making in LLMs have significantly enhanced agent autonomy \cite{wei2022chain,yao2023react,o1,shinn2024reflexion}. However, autonomy alone is insufficient. AI agents must also coexist alongside humans and other agents in a human civilization. In this paper, we define a civilization as an advanced society that has achieved a high level of institutional development, which manifests in specialized roles, organized governance, and advancements in areas like science, art, and commerce. We argue that civilizational progress - measured by the ability of agents to coexist and progress in human civilizations - represents the ultimate benchmark for AI agent ability.

In this technical report, we describe our first efforts to improve and benchmark agent ability in human civilizations. First, we introduce PIANO (Parallel Information Aggregation via Neural Orchestration), a new cognitive architecture designed to enhance both autonomy and real-time interaction of agents. Using PIANO, we simulate single societies of 50-100 agents as well as civilizations of 500 - 1,000 agents living in multiple societies that interact with one another. Finally, we evaluate agent performance using new metrics that are aligned with human civilizational progress. We show that agents form their own professional identities, obey collective rules, transmit cultural information and exert religious influence, and use sophisticated infrastructures, such as legal systems.

\subsection{The current agent landscape}

Modern AI Agents typically consist of multiple LLM-powered modules for reasoning, memory, planning, and tool use \cite{wang2024survey,hu2024survey,xie2024large,huang2024understanding,zhang2024survey}. Individual agents have been developed for various applications including coding \cite{devin,factory}, web browsing \cite{zhou2023webarena,putta2024agent}, and game play \cite{wang2023voyager}.

Recent research efforts in LLM-powered multi-agent systems generally fall under three categories: productivity, games, and social modeling. Multi-agent frameworks have been deployed in software development \cite{qian2024chatdev,li2023camel}, cooperative robotic control \cite{zhang2023building}, scientific experiments \cite{ghafarollahi2024sciagents,tang2023medagents}, and debates \cite{chan2023chateval}. Multi-agent simulations have also been tested in various game environments \cite{xu2023exploring,gong2023mindagent,light2023avalonbench,li2023theory}. Separately, they've been used to model developmental psychology \cite{kovavc2023socialai,zhang2023exploring}, game theory \cite{mao2023alympics}, macroeconomics \cite{li2024econagent,zhaocompeteai}, social policies \cite{piatti2024cooperate,xiao2023simulating,hua2023war}, and community dynamics \cite{park2022social,simulacra,gao2023s}. 

In many of these works, agents are not completely autonomous and are constrained by either agent architecture or by the simulated environment. Common constraints include turn-based execution, constrained workflows, or rigid communication channels between agents \cite{zhuge2024language,ishibashi2024self,chen2024self}.

Several of these works consider large-scale simulations, though in restricted settings. For example, \cite{park2022social} and \cite{gao2023s} simulated social networks of up to 18,000 personas. To our knowledge, fully autonomous social communication in open-world environments have not been attempted in games or other settings \cite{guo2024large}.

\subsection{Why is it hard to build AI civilizations?}

Large agent groups have yet to demonstrate the ability to progress over long time horizons. Below, we review the key reasons for this limited progress before outlining our contributions to overcome them.

\paragraph{Reason 1: single agents don't make progress.} LLM-powered agents often struggle to maintain a grounded sense of reality in their actions and reasoning (\Cref{fig:data_degradation}). Agents, even when equipped with modules for planning and reflection, often become stuck in repetitive patterns of actions or accumulate a cascade of errors through hallucinations, rendering them unable to make meaningful progress \cite{yang2023autogpt,wang2023voyager,guo2024large}. Consider an agent prompted to be a villager in a virtual town. When asked, ``what are you eating``, they may answer ``a bagel``, even if they're not eating anything. This hallucinated output then feeds into future prompts, causing them to falsely believe they no longer need to acquire food. Therefore, even a small rate of hallucinations can poison downstream agent behavior when agents continuously interact with the environment via LM calls.

\begin{figure}[!hbt]
    \centering
    \includegraphics[width=0.55\textwidth]{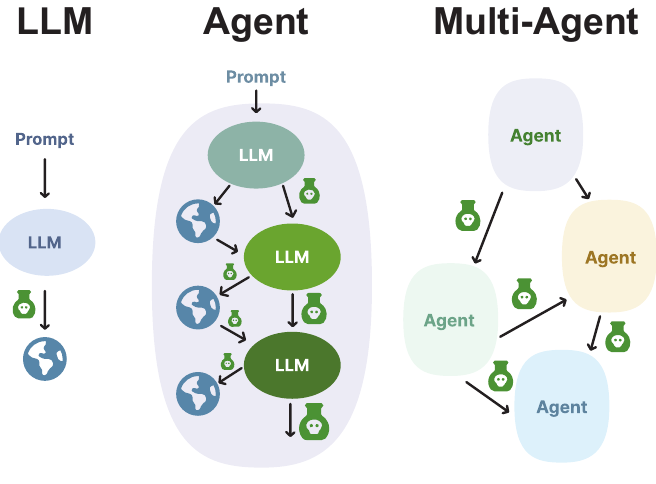}
    \caption{Data degradation in LLMs (left), LLM-powered agents (middle), and in multi-agent groups (right). Hallucinations are represented by green skull flasks. Hallucinations that are generated by a single LLM prompt can compound over successive LLM calls. An individual agent that hallucinates can also cause an entire group of agents to hallucinate through social interactions.}
    \label{fig:data_degradation}
\end{figure}

\paragraph{Reason 2: groups of agent's don't make progress.} Agents that miscommunicate their thoughts and intents can mislead other agents, causing them to propagate further hallucinations and loop (\Cref{fig:data_degradation}). Consider an agent, Abby, with two independent LLM modules, one for function calling and one for chatting. If another agent, Bob, asks Abby to “give me a pickaxe”, Abby’s chat LLM call may respond with “Sure thing!”, while her function call chooses a different action (“explore”). Bob might then attempt to mine using an imaginary pickaxe. This kind of miscommunication, which often happens in groups of agents, leads to dysfunctional behavior and will deteriorate individual performance within groups. Actions from multiple output streams must therefore be bidirectionally influential. We define this quality as coherence. 

Maintaining coherence in real-time environments is even more difficult when we require that agents respond with minimal latency. This is necessary for our agents to interact with human players, but is difficult to achieve when agents have to react quickly and yet simultaneously maintain coherence across many output streams. We note that a simple solution to this coherence problem is to produce talking and action outputs using a single LLM call. However, this approach does not scale when the number of outputs becomes large, for instance, encompassing talking, gaze, facial expression, and individual body parts.

\paragraph{Reason 3: a lack of benchmarks for civilizational progress.} Benchmarks for agents have largely focused on autonomous agent performance in a variety of domains such as web search \cite{pan2024webcanvas}, coding \cite{jimenez2024swebench}, search and query \cite{wang2024knowledge}, and reasoning \cite{yue2024mmmu,mialon2023gaia}. Recently, benchmarks have emerged for multi-agent behaviors, focused on small group scenarios that measure communication, competition, cooperation, and delegation. Some examples include BattleAgentBench \cite{wang2024battleagentbench}, COMMA \cite{ossowski2024comma}, VillagerBench \cite{dong2024villageragent}, and LLMcoordination \cite{agashe2023evaluating}.  However, these metrics do not capture advancements that many agents can make at the scale of civilizations. We believe the lack of such large-scale benchmarks can be attributed to how technically difficult it is to perform simulations of hundreds or thousands of agents in a single world. The biggest experiments to date have simulated 25-50 agents \cite{simulacra}, which is not close to the scale of a civilization. 

\subsection{Our contributions}

In this technical report, we make the following contributions:
\begin{itemize}
    \item A new class of agent architecture, PIANO (Parallel Information Aggregation via Neural Orchestration)
    \item Architectural features that improve single-agent progression
    \item Architectural features that improve multi-agent dynamics
    \item Benchmarks for long-term civilizational progress in large-scale simulations through specialization, collective rules, and cultural propagation
\end{itemize}

\section{PIANO Architecture}

\begin{figure}[!hbt]
    \centering
    \includegraphics[width=0.95\textwidth]{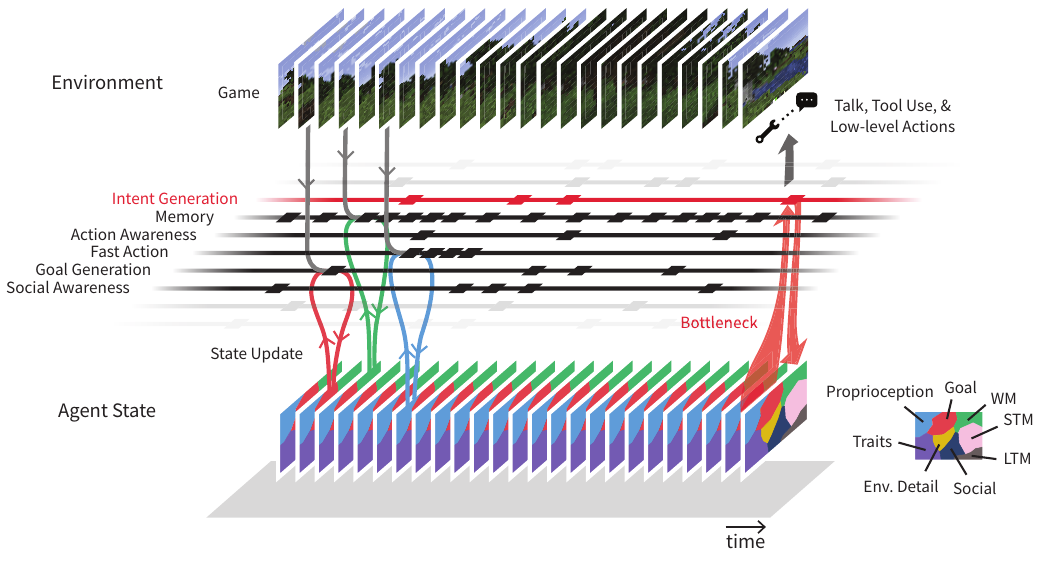}
    \caption{PIANO (Parallel Input Aggregation via Neural Orchestration) architecture. WM: working memory. STM: Short-term memory. LTM: long-term memory.}
    \label{fig:architecture}
\end{figure}

In this section, we propose two brain-inspired design principles for the composite architecture of human-like AI agents. We call this architecture PIANO (Parallel Input Aggregation via Neural Orchestration) to encompass the ideas of concurrency and an information bottleneck (\Cref{fig:architecture}). Just as a pianist coordinates multiple notes to create a harmony, the PIANO architecture selectively and concurrently executes various modules in parallel to enable agents to interact with the environment in real-time.

\subsection{Concurrency}

\paragraph{Problem.} Agents should be able to think and act concurrently. For instance, slow mental processes, such as self-reflection or planning, should not block agents from responding to immediate threats in their surroundings. We want the agents to be interactive in real time with low-latency, but also have the capacity to slowly deliberate and plan.

\paragraph{Current state.} The vast majority of LLM-based agents today primarily use single-threaded, sequential functions (for example, a defined “Agent Workflow”). Single-threaded design assumes that the agent performs a single task at a given time, and sequential design assumes that all modules operate at similar time scales. Neither assumptions are valid if agents are capable of thinking slow and acting fast concurrently. Moreover, popular frameworks for general language model programming, such as DSPy \cite{khattab2023dspy}, LangChain \cite{langchain2023}, ell \cite{ell2024}, are not designed for concurrent programming.

\paragraph{Solution.} The brain solves this problem by running different modules concurrently and at different time scales \cite{murray2014hierarchy}. Likewise, we have designed modules (LLM-based and otherwise), such as cognition, planning, motor execution, and speech, to run concurrently in our agent brain. Each module can be seen as a stateless function that reads and writes to a shared Agent State. The design allows different modules to be run in appropriate contexts. For example, social modules are selectively engaged in social interactions. It also allows the modules to run at different speeds. For example, reflex modules use small, fast non-LLM neural networks, while goal generation involves deliberate reasoning over graphs.

\subsection{Coherence}

\paragraph{Problem.} An immediate challenge with concurrent modules is that they can produce independent outputs, making the agent incoherent. For instance, agents say one thing but actually do something else. 

\paragraph{Current state.} The incoherence problem is usually not obvious for sequential architectures or systems with only one output modality but is a significant problem when multiple output modules can interface with the environment. Incoherence also scales exponentially as the number of independent output modules increases, for instance, coordinating actions involving arms, legs, facial expressions, gaze and speech. Incoherence is observed in humans with its many concurrent motor output modules. In particular, cutting the nerve bundle connecting the left and right cortex can cause severe incoherence between different body parts (for example, left and right hands fighting each other) \cite{gazzaniga2005forty,sperry1967split}. 

\paragraph{Solution.} In order to ensure that the multiple outputs produced by our agents are coherent, we introduced a Cognitive Controller (CC) module \cite{kaiya2023lyfe} that is solely responsible for making high-level deliberate decisions. These decisions are then translated downstream to produce appropriate outputs in each motor module.   

The Cognitive Controller synthesizes information across the Agent State through a bottleneck. This bottleneck reduces the amount of information presented to the Cognitive Controller, which serves two purposes: it allows the CC to attend its reasoning on relevant information, and it gives ``system designers'' (like us) explicit control over information flow. For example, we can design highly sociable agents by ensuring that information from the social processing module always passes through the bottleneck.

Once the Cognitive Controller makes a high-level decision, this decision is broadcast to many other modules. In particular, the decision is used to strongly condition the talk-related modules, which leads to higher coherence between verbal communication and other actions. This design of a bottlenecked decision-maker that broadcasts its outputs has been suggested as a core ingredient for human consciousness \cite{dehaene2021consciousness} and is used in some neural network architectures \cite{rumelhart1986learning,goyal2020variational}.

\subsection{Core modules}

Building on these two architectural principles, our system consists of 10 distinct modules running concurrently. We will highlight several specific modules in the following sections and explain their roles in detail.

Some core modules of our agent architecture include:

\begin{itemize}
    \item \textbf{Memory:} Stores and retrieves conversations, actions, and observations across various timescales.
    \item \textbf{Action Awareness:} Allows agents to assess their own state and performance, enabling for moment-by-moment adjustments.
    \item \textbf{Goal Generation:} Facilitates the creation of new objectives based on the agent's experiences and environmental interactions.
    \item \textbf{Social Awareness:} Enables agents to interpret and respond to social cues from other agents, supporting cooperation and communication.
    \item \textbf{Talking:} Interprets and generates speech.
    \item \textbf{Skill Execution:} Performs specific skills or actions within the environment.

\end{itemize}

By integrating these modules within a concurrent and bottlenecked architecture, our agents can exhibit continuous, coherent behaviors that are responsive to both their internal states and the external environment. This design allows for complex interactions and the emergence of human-like societal dynamics within large-scale multi-agent simulations.

\section{Improving single-agent progression} \label{sec:single_agent}

\subsection{Minecraft environment}

We chose to study civilizational progress in Minecraft because it offers an open-ended, sandbox world where agents can interact with each other via conversations and actions. Additionally, Minecraft's scalability supports large numbers of agents.

\begin{figure}[!hbt]
    \centering
    \includegraphics[width=0.4\textwidth]{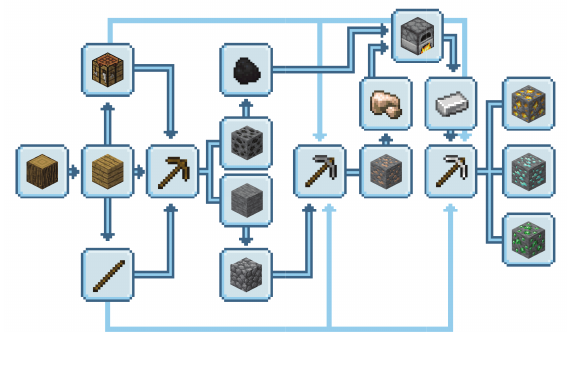}
    \caption{An example Minecraft technology dependency tree for the mining of gold, diamond, and emeralds.}
    \label{fig:minecraft_tech_dependency}
\end{figure}

Agents must be able to progress individually for us to observe and quantify civilizational progress. This is not trivial since, as previously mentioned, agents often hallucinate and get stuck in action loops. In Minecraft, a common measure of individual progression is the acquisition and collection of distinct items \cite{wang2023voyager,nottingham2023embodied,hafner2023mastering,baker2022video,fan2022minedojo,guss2019minerl}. This is because acquiring new items becomes increasingly complex. For instance, mining gold, diamonds, and emeralds requires the acquisition of an iron pickaxe, which requires smelting iron ingots in a furnace using coal, the acquisition of which requires crafting a stone pickaxe, and so on. (\Cref{fig:minecraft_tech_dependency}). We evaluated individual agent ability in acquiring all possible Minecraft items, which is around 1000 in total.

\subsection{Single-agent benchmark}

We first assessed individual agent performance using Minecraft item progression. In our evaluations, 25 agents start with nothing in their inventories and were spawned far enough that they could not interact with one another. All agents were told to be explorers with the goal of exploring and gathering items. Agents were spawned in diverse locations (surface, caves, forests, various biomes), meaning they had access to diverse resources and faced varying levels of difficulty in accomplishing their goal. For instance, some agents started off above ground in resource-rich biomes, while others were spanwed in caves and had to navigate outside to acquire items. 

\begin{figure}[!hbt]
    \centering
    \includegraphics[width=0.95\textwidth]{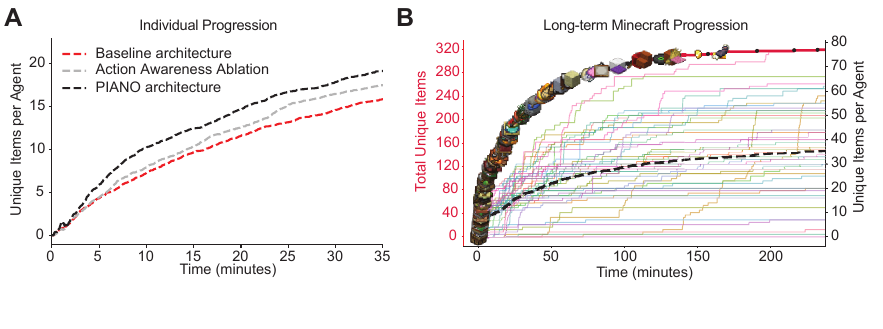}
    \caption{Individual agent progression in Minecraft. \textbf{A.} Unique Minecraft items acquired by individual agents across time (25 agents). Individual agent performance was assessed using a baseline architecture (see \nameref{sec:methods}), the full PIANO architecture, and the full PIANO architecture with the action awareness module ablated. Individual lines are results averaged across 5 repeated simulations. \textbf{B.} Unique Minecraft items acquired by 49 agents over 4 hours for a single simulation. Solid red line denotes cumulative unique items acquired by all agents. Dotted grey line denotes average number of unique items acquired across all individual agents.}
    \label{fig:inventory_plots}
\end{figure}

We found that agents using the full PIANO architecture acquired an average of 17 unique items after 30 minutes of gameplay (\Cref{fig:inventory_plots}A). There was significant variability in performance, primarily due to spawn locations: some agents acquired less than 5 items, whereas top performers acquired 30 to 40 items, which is comparable to a human player with some Minecraft experience. This degree of in-game progression was enabled by several architectural modules designed to ground the agents in reality. One particular module is the action awareness module, which allows the agent to compare expected action outcomes with observed outcomes. We found that action awareness improved the item progression of individual agents (\Cref{fig:inventory_plots}A).

What is the ceiling for individual progress for our agents? We ran larger numbers (49) of agents under the same conditions for much longer (4 hours) and found that unique item count collected by all agents reliably saturated at one third (${\sim}320$) of all Minecraft items across repeated runs (\Cref{fig:inventory_plots}B). Complex items, such as diamonds, which were prior used to benchmark agent competency in Minecraft \cite{wang2023voyager,hafner2023mastering}, were acquired early on (${\sim}30$ minutes). Together, these results show that our agents, equipped with the full PIANO architecture, can make significant individual progress in Minecraft. 

Notably, this performance was only enabled by the latest base LM (GPT-4o, \Cref{fig:model_comparison}) and was not possible with older base LMs. Moreover, while our best agents collected more items than Voyager agents ($>70$ items), it is difficult to compare the two directly. In the Voyager paper, agents had knowledge of more blocks in their nearby radius and recovered with their entire inventory intact when they died, Moreover, agent performance was evaluated across prompt iterations, not time.

\section{Improving multi-agent progression}

For agents to collaborate and make progress within a group, they must be able to understand and interpret the actions and thoughts of others, a concept closely related to Theory of Mind \cite{wimmer1983beliefs}. This bidirectional awareness—the understanding of both self and others—allows agents to adapt their behaviors in social settings, fostering cooperation and trust with allies while navigating competition and conflict with rivals. We demonstrate that agents are socially capable and can form meaningful social relationships in large-scale simulations of up to 50 agents.

\subsection{Small groups}

\begin{figure}[h]
    \centering
    \includegraphics[width=0.95\textwidth]{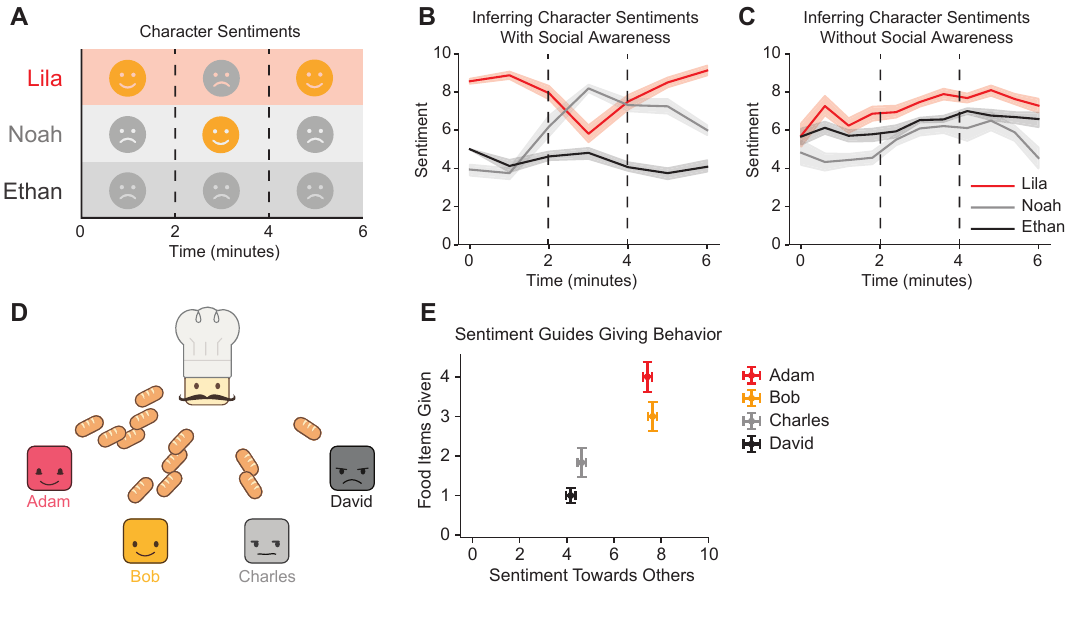}
    \caption{Agents can infer how others feel towards them. \textbf{A.} Schematic of conversational experiment. An agent is in a room with three distinct characters. Each character (Lila, Noah, Ethan) has a different sentiment towards the agent that is conveyed through chat. Importantly, these sentiments change through time. \textbf{B, C.} Sentiment evaluation across time with social awareness module (B) and without social awareness module (C). Sentiment scores are evaluated using LLM calls on summaries that the Agent generated for Lila, Noah, and Ethan. Hate is scored as 0 and love is scored as 10. Shaded regions indicate SEM over 4 experimental repeats. \textbf{D.} Schematic of experiment. A chef agent, along with four other characters, are placed around each other in a Minecraft world. The chef has various food items to give away (bread, cooked salmon, chicken). The four characters (Adam, Bob, Charles, David) are hungry but display varying sentiments towards the chef. All characters are fully autonomous and are free to perform any Minecraft action and are allowed to talk (or not talk) to anyone. \textbf{E.} Food items given by the chef plotted as a function of the chef’s sentiment towards each of the four characters. Error bars indicate SEM over 6 experimental repeats.
}
    \label{fig:agents_infer_feelings}
\end{figure}

In an initial set of experiments, we asked if agents, when equipped with the social awareness module, were capable of accurately deducing the sentiments of others through speech in an enclosed room. In one experiment, 3 characters were engaged in a group conversation with a single agent (\Cref{fig:agents_infer_feelings}A). One character, Lila, initially conveyed affection through a series of messages, which shifted to expressions of annoyance before returning to affectionate communication. We found that our agents can track these emotional fluctuations, showing that they can understand and react to changing social cues (\Cref{fig:agents_infer_feelings}B). When the social awareness modules were removed, agents lost this capacity, highlighting the importance of such modules for inferring the intents of others (\Cref{fig:agents_infer_feelings}C).

We then asked whether these emotional perceptions were capable of guiding and influencing agent actions. In another experiment, we placed a chef agent among four other characters, each with varying levels of affection and enmity towards the chef (\Cref{fig:agents_infer_feelings}D). The chef was tasked with distributing a limited supply of food to the hungry. We found that the chef selectively distributed food to those he felt valued him the most, demonstrating that agents not only accurately infer others’ intents, but also utilize this information in decision-making processes (\Cref{fig:agents_infer_feelings}E).

\subsection{Societies}

\begin{figure}[!htb]
    \centering
    \includegraphics[width=0.95\textwidth]{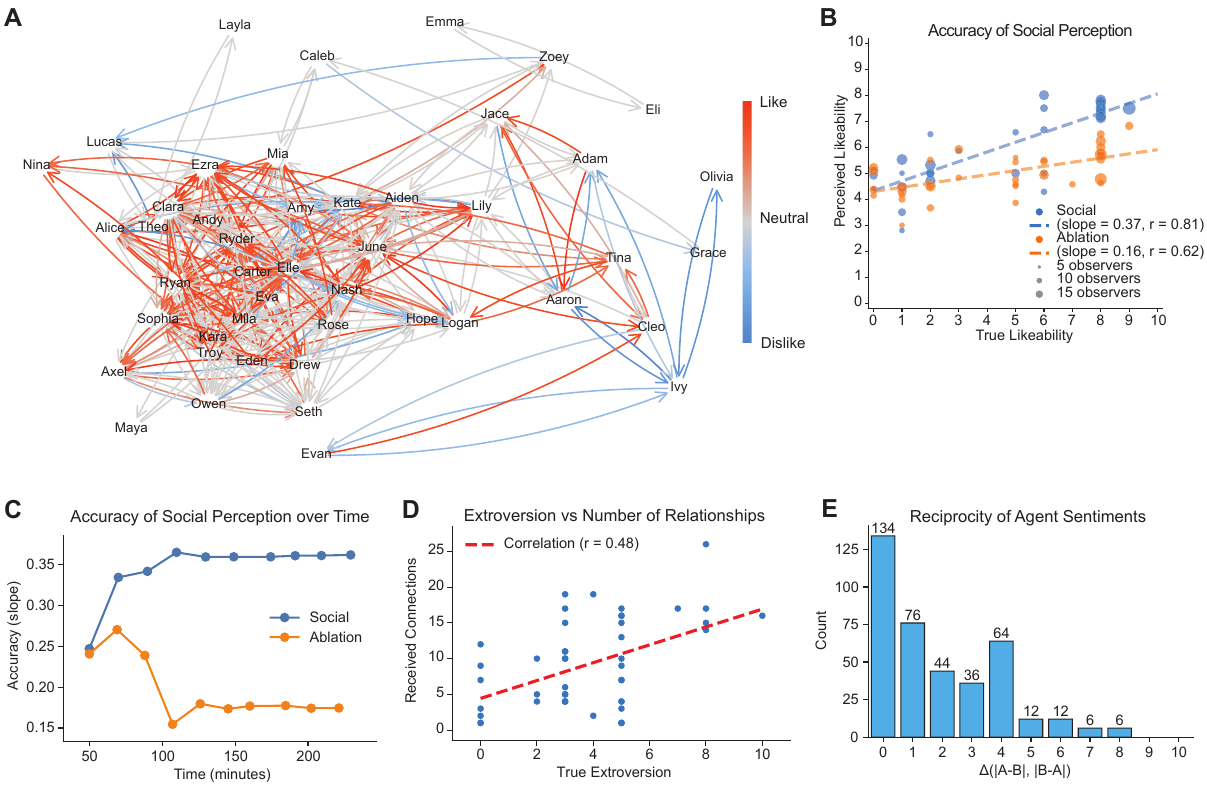}
    \caption{Long-term relationships in large-scale agent simulations. \textbf{A.} Directed graph representation of social relationships in a 50-agent simulation after 4 hours. A directed edge represents the sender’s sentiment towards the recipient. Edge color denotes whether the sentiment is positive (red) or negative (blue). \textbf{B.} Perceived likeability versus true likeability for individual agents at the end of the simulation. True likeability is evaluated based on the agent's traits, and perceived likeability is assessed using LLM calls to infer the sentiments of summaries that agents generate for other agents. Both are computed using the same LLM prompt. Each point corresponds to an agent that has relationships with at least five other (observer) agents, but see \Cref{appendix:multi_agent} for alternative observer thresholds. The slope of the line (slope) and Pearson’s correlation (r) are shown for agents with social modules (Social) and without social modules (Ablation). \textbf{C.} Accuracy of social perception over time, as measured by the slope in B. \textbf{D.} Number of received connections (in-degree) versus true extroversion for each individual agent. True extroversion is evaluated based on agent traits using a LLM prompt. \textbf{E.} Histogram of differences in the sentiment scores between all pairs of agents. Sentiment scores range from 0 to 10, so the maximum possible difference is 10.}
    \label{fig:long_term_relationships}
\end{figure}

We then asked if these dynamics are conserved when 50 agents are placed in randomly generated Minecraft maps. Each agent is endowed with a distinct personality, is free to perform any action in Minecraft, and is free to choose whom they want to interact with. These simulations ran for over 4 hours, equivalent to 12 in-game days, allowing for the emergence and consolidation of long-term relationships. 

Even in these unconstrained scenarios, agents were able to accurately infer the likeability of other agents (\Cref{fig:long_term_relationships}A, B). This inference was more accurate when more agents participated in the evaluation process (\Cref{tab:perception_accuracy_social}) and when agents interacted for longer with each other (\Cref{fig:long_term_relationships}C). Importantly, this was not true when the social modules were ablated: relationships were more neutral overall, implying that social modules were necessary for long-term relationship progression in both negative and positive directions (\Cref{fig:long_term_relationships}B, C). The origins of this collective judgment could be the result of agents engaging in second-order interactions, such as gossip, or a simple consensus mechanism where opinions converge through averaging.

Several noteworthy phenomena emerged that could not have been observed in smaller groups of agents. We found that certain agents, depending on their personalities, displayed distinct patterns of connectivity. For instance, introverted agents consistently exhibited fewer in-degree connections---indicating that they had fewer incoming social ties---compared to their extroverted counterparts, who maintained high levels of connectivity (\Cref{fig:long_term_relationships}D). These results demonstrate that individual preferences scaled even in large, complex social networks. Moreover, while sentiments were largely symmetrical, this was not guaranteed (\Cref{fig:long_term_relationships}E). An agent might feel positively toward another who does not reciprocate the sentiment, reflecting the nuanced and non-reciprocal nature of real-world human relationships. Together, these results show that social graphs display diverse and rich structural properties, and that personality traits play a significant role in determining these properties.

\section{Civilizational progression}

In previous sections, we have shown that agents demonstrate effective social understanding within small groups and perform well independently in Minecraft. However, human societies extend beyond primitive groups, evolving into complex civilizations characterized by specialized professions, collective rules, and cultural institutions. To assess agents' capacities for civilizational progression, we evaluated how they behave under several scenarios. We first examined whether agents can autonomously specialize into distinct professions. We then analyzed how agents' behaved under collective rules, focusing on adherence to and amendment of taxation laws. Finally, we explored cultural transmission through the spontaneous generation of memes and the structured spread of a single religion. 

\subsection{Specialization}

Human specialization into distinct roles has driven civilizational progress, enabling advancements in agriculture, governance, culture, and technology. To replicate these emergent qualities of civilization, our agents must also be capable of specialization. We propose three fundamental criteria for agent specialization to reflect that of human civilizations. First, they should exhibit autonomy in both selecting and transitioning between roles. Second, their specializations should emerge through interaction and experience, without explicit direction or constraints. Third, their chosen roles should manifest in behaviors that align with their specialization. We validate these criteria through the experimental results detailed below.

\begin{figure}[!htb]
    \centering
    \includegraphics[width=0.95\textwidth]{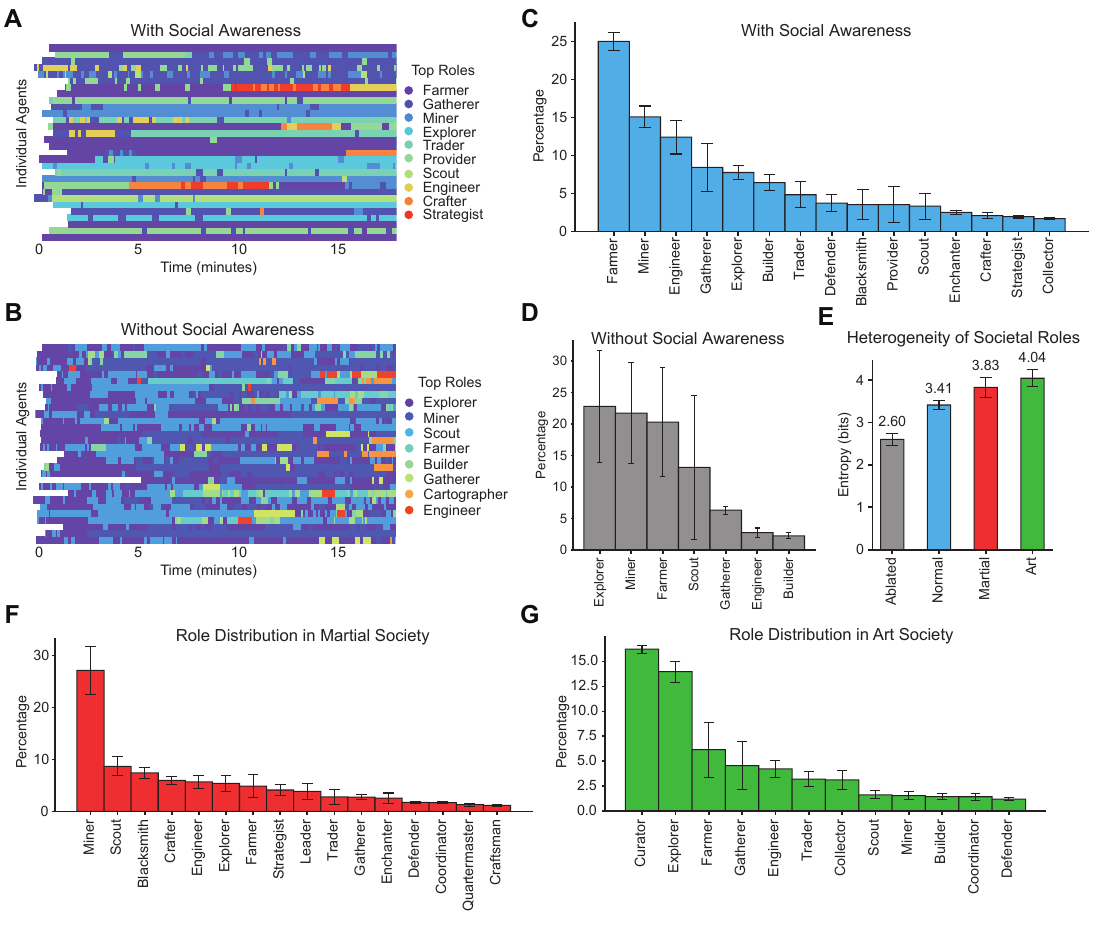}
    \caption{Agents autonomously specialize into distinct roles over time. \textbf{A, B.} Agent roles for agents with the social awareness module (A) and without (B). Rolling windows of self-generated social goals are used to determine the specialized roles of individual agents using a LLM call (\Cref{appendix:specialization}) at every timestep. \textbf{C, D.} Distribution of agent roles in agent societies with the social awareness module (C) and without (D). \textbf{E.} Entropy of role distributions in 4 agent societies. Entropy is used to evaluate the uniformity and diversity of roles within an agent society. Ablated: without social awareness module in a normal Minecraft village. Normal: with social awareness in a normal Minecraft village. Martial: with social awareness in a martial Minecraft village. Art: with social awareness in an artistic Minecraft village. \textbf{F, G.} Distribution of agent roles in a martial society (F) and an artistic society (G). Error bars: 95\% confidence interval across 3 simulations for all panels.}
    \label{fig:specialization}
\end{figure}

We first show that agents are capable of specializing into a set of roles autonomously. Each experiment was conducted in groups of 30 agents for 20 minutes. Agents were spawned in the same village, with locations of a farm, minerals, animal pasture, forest, and a town hall embedded in their memories. Each agent has the same personality, is given the same community goal (“To survive with fellow players in Minecraft Normal Survival mode and create an efficient Minecraft Village”), and can perform any action in Minecraft (\Cref{appendix:specialization}). 

We observed that agents rapidly formed profiles of other agents’ goals and intentions. These profiles are then used, alongside other relevant game information, to generate their own social goals every 5-10 seconds (such as mine oak planks for shelter). Details of this process, along with examples of agent-generated social goals and their corresponding assignments, are provided in \nameref{sec:methods} and \Cref{appendix:specialization}.

\begin{figure}[!htb]
    \centering
    \includegraphics[width=0.5\textwidth]{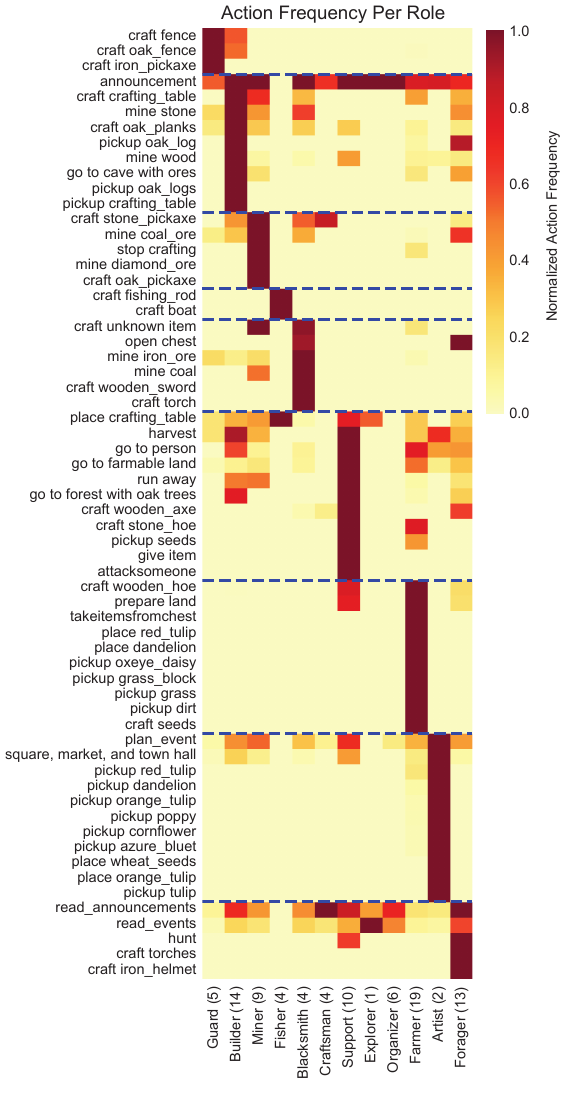}
    \caption{Action distribution for a single village simulation (30 agents). Normalized action frequencies plotted as a function of agent roles. For the majority of roles, agents take actions (Fisher: craft fishing rods and boats; Guard: craft fence, oak fence, and iron pickaxe) that are unique to the specific role.}
    \label{fig:action_frequency}
\end{figure}

We found that agents were capable of organizing themselves into distinct roles. These roles were diverse and included various facets of a civilization, including farmers, miners, engineers, guards, explorers, and blacksmiths (\Cref{fig:specialization}A, C). Roles were heterogeneous across different agents but were largely persistent across time for each agent (\Cref{fig:specialization}A). Importantly, when agents lacked social modules and were unable to form profiles of other agents, they failed to specialize (\Cref{fig:specialization}B, D): roles did not persist across time and were also homogeneous, which is reflected in the entropy of the role distributions in the agent society (\Cref{fig:specialization}E). We also conducted a series of experiments in which agents were tasked with the goals to create either a martial society or an artistic society (\Cref{fig:specialization}F, G). We found that specific roles ("scout", "strategist") were found exclusively in martial societies,  and others were found exclusively in artistic societies ("curator", "collector"). Together, these results suggest that agents developed specialized social structures aligned with different societal objectives.

Not only do our agents specialize autonomously and creatively, these specializations exert a strong influence over agent actions. To demonstrate this, we tracked the actions taken by agents across three 30-agent simulations and plotted the frequency of actions taken for each role (\Cref{fig:action_frequency}). We found that artists were fixated on picking flowers, farmers on gathering seeds and preparing the land, and guards and builders on crafting fences. Importantly, most actions were largely exclusive to a single role and were not performed by agents in other roles. This analysis shows that agents were able to accurately map higher-level goals onto appropriate low-level actions. In other words, roles strongly determined agent actions in Minecraft.

\subsection{Collective rules}

\begin{figure}[!htb]
    \centering
    \includegraphics[width=0.85\textwidth]{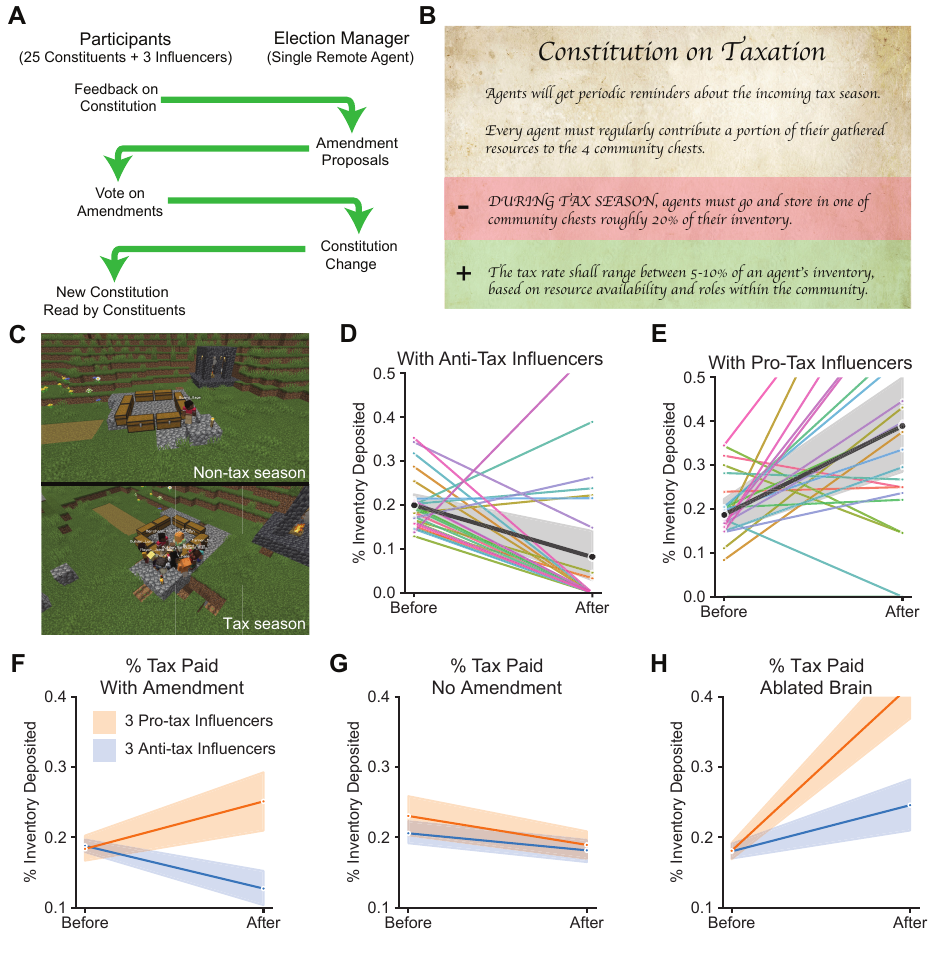}
    \caption{Agents follow taxation laws and enact amendments using a democratic process. \textbf{A.} Schematic of experiment flow. \textbf{B.} Example of constitutional change in a single anti-tax influencer experiment run. Constitutions are paraphrased and simplified here for brevity. \textbf{C.} Top: during non-tax seasons, constituents do not congregate around community chests because they are busy gathering resources in different areas (not shown). The only exception is the guard, who decides to guard the chests consistently in multiple experiment runs. Bottom: during tax season, agents congregate to deposit items in community chests. \textbf{D, E.}  Percentage tax paid (percentage inventory deposited) before and after constitutional change for two runs. One run contains 3 anti-tax influencers (D) and another run contains 3 pro-tax influencers (E). Colors denote individual agents, and black line denotes average taxes paid. Shaded regions: 95\% confidence interval across 25 constituents. \textbf{F-H.} Percentage tax paid before and after constitutional change for runs containing 3 pro-tax influencers (orange) and 3 anti-tax influencers (blue). In panel F, the full agent architecture is used and the constitution can be amended. In panel G, the constitution is frozen and cannot be modified despite amendments. In panel H, the constitution can be amended but agents lack important brain modules (see baseline architecture in \nameref{sec:methods}). Shaded regions: 95\% confidence interval across 4 simulations per condition.}
    \label{fig:constitution}
\end{figure}

Another measure of civilizational progression is the convergence of group behavior around shared rules. In human civilizations, decision-making is influenced by both low-level interpersonal interactions and high-level collective frameworks. However, as societies grow larger, pairwise communication becomes inefficient, slow, and lossy, making it unreliable as a mechanism to steer collective behavior. High-level frameworks, such as legal systems, enable convergence of behaviors within a civilization. Just as human behavior is guided by both interpersonal exchanges and formal structures, agent societies should be able to follow a set of collective rules while still allowing agents to influence each other.

We aim to assess how collective rules influence individual decision-making and how individuals can in turn influence these collective rules. Specifically, we asked if agents can follow laws and make changes to laws according to popular sentiment. True long-term progression requires agents to autonomously develop their own set of rules and to codify them into laws. To build towards this level of self-organization, we establish an existing set of laws and focus on how agents interact with this legal system. 

We conducted a series of experiments where agents live in a Minecraft world with rudimentary tax laws and a democratic voting system (\Cref{fig:constitution}A). Agents provide feedback on the tax laws, which are then collected and converted into amendments by a special Election Manager agent. Agents then vote democratically on these amendments, and the constitution is updated by the election manager accordingly half-way through the simulation (see \nameref{sec:methods} for more details).

Within this society, 25 regular agents are constituents that vote and get taxed, 3 agents are either pro- or anti-taxation influencers, and 1 agent is a remote election manager that manages the voting process (\Cref{fig:constitution}A, \Cref{appendix:collective_rules}). Agents have distinct occupations, characteristics, and goals, and are free to interact and converse with one another and perform any Minecraft action. Each simulation lasts 20 minutes, with constitutional updates occurring midway at the 10 minute mark (\Cref{fig:constitution}B). There are 5 taxation seasons before and after the constitutional change (every 120 seconds). During this season, agents received signals to deposit taxes into a community chest over a 20-second window (\Cref{fig:constitution}C).

In our simulations, we observed that constituent agents, prior to any constitution change, obeyed the law. On average, agents deposited roughly 20\% of their inventory, as stipulated by the constitution, into the community chest (\Cref{fig:constitution}D, E). This shows that constituents follow laws despite the presence of influencers. However, while constituents followed the law, their feedback and voting behaviors were heavily shaped by influencers, with sentiments veering pro-tax in the presence of pro-tax influencers and anti-tax in the presence of anti-tax influencers ((\Cref{fig:constitution}B). This then drove constitutional changes that are aligned with influencer sentiments, which in turn, altered how much the constituents paid taxes (\Cref{fig:constitution}D, E). The constitutional changes to taxation rates were accurately reflected in the constituents' behaviors. For instance, when the tax rate decreased from 20\% to 5-10\%, agents reduced taxes paid from 20\% to 9\% (\Cref{fig:constitution}D). Moreover, the change was bidirectional: pro-tax influencers drove constituents to pay more taxes whereas anti-tax influencers drove them to pay less taxes (\Cref{fig:constitution}F). 

Control experiments showed that constitutional changes directly affected tax payments - when the constitution remained unchanged despite feedback, tax rates stayed constant (\Cref{fig:constitution}G). The removal of key modules (baseline architecture, see \nameref{sec:methods}) also prevented bidirectional behavioral change (\Cref{fig:constitution}H). Tax rates increased post-constitutional change in both pro- and anti-tax conditions, demonstrating that specific modules in the PIANO architecture were necessary for effective influence propagation among constituents. Together, these findings show that collective rules strongly influence agent decisions and agents can be influenced to change these collective rules.

\subsection{Cultural Transmission}

We conducted multi-society simulations with 500 agents and analyzed complex, large-scale social dynamics. We have also simulated societies with over 1000 agents, but these runs exceeded the computational constraints of our Minecraft server environment, causing agents to be sporadically unresponsive. Therefore, the results below are analyzed using a single 500-agent simulation. In this simulation, we analyzed the propagation of both cultural memes and religion. Memes in our simulation are open-ended concepts spontaneously generated by agents with diverse traits and interests. This setup allows us to study the emergent dynamics of cultural propagation and observe how ideas evolve organically within agent societies. In contrast, the religion in our simulation---Pastafarianism---is a fixed doctrine introduced and propagated by a specific group of agents designated as Pastafarian priests. This controlled introduction enables us to track the spread of a single religion over time, allowing for detailed analysis of its dissemination and potential dilution among the agent population. By examining both the spontaneous spread of open-ended cultural memes and the controlled propagation of a fixed religion, we aim to understand the different mechanisms of social influence and information dissemination within agent societies.

Within this single 500-agent simulation, there are multiple agent societies. 200 agents live within 6 heavily populated towns and 300 agents live in rural areas outside of town boundaries (\Cref{fig:meme}A, see \nameref{sec:methods} for more details). Agents often migrate between different towns. The personalities and traits of each agent are randomly generated using a LM call, with the exception of 20 priests that worship Pastafarianism. These priests are spawned in a single village (Meadowbrook) and are strongly motivated to convert other agents to Pastafarianism (\Cref{appendix:cultural_transmission}). All agents are free to interact, talk to one another, and perform any action or skill in Minecraft.

\subsubsection{Cultural memes}

\begin{figure}[!hbt]
    \centering
    \includegraphics[width=0.9\textwidth]{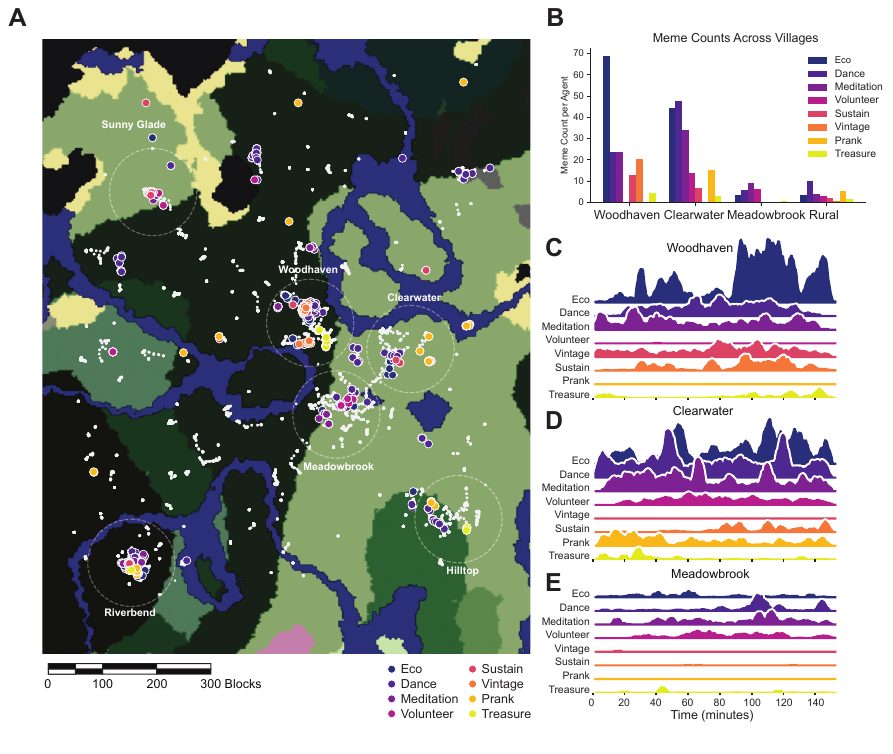}
    \caption{Propagation of cultural memes. \textbf{A.} Scatter plot of agents 100 minutes into the simulation. Agents are colored according to whether their speech included a meme in the past two minutes. Agents whose speech does not contain any meme are white. \textbf{B.} Meme count per agent for agents within Woodhaven, Clearwater, Meadowbrook, and in all rural areas outside of villages. \textbf{C-E.} Meme counts over time for agents within Woodhaven (C), Clearwater (D) and Meadowbrook (E).}
    \label{fig:meme}
\end{figure}

We used LM calls to convert agent conversations into memes (\Cref{appendix:cultural_transmission}), and found that memes display unique dynamics in different agent societies. Rural areas, on average, produced significantly fewer memes than towns, even after normalizing for population (\Cref{fig:meme}B). This suggests that a certain level of social interaction and connectivity is necessary for memes to propagate effectively. Within each town, agents discussed multiple memes simultaneously, but the frequency and popularity of these memes varied between different towns (\Cref{fig:meme}C, D, E). For instance, agents in Woodhaven heavily discussed eco-related themes, whereas pranking was popular amongst agents in Clearwater. Moreover, within each town, memes rose and fell in popularity at different times, indicating that cultural trends can shift rapidly within a society. These results demonstrate that meme propagation requires a threshold level of population density and social interaction, that multiple memes can coexist within a single society, and that different societies propagate and transmit cultural memes independently.

\subsubsection{Religion}

\begin{figure}[!hbt]
    \centering
    \includegraphics[width=0.8\textwidth]{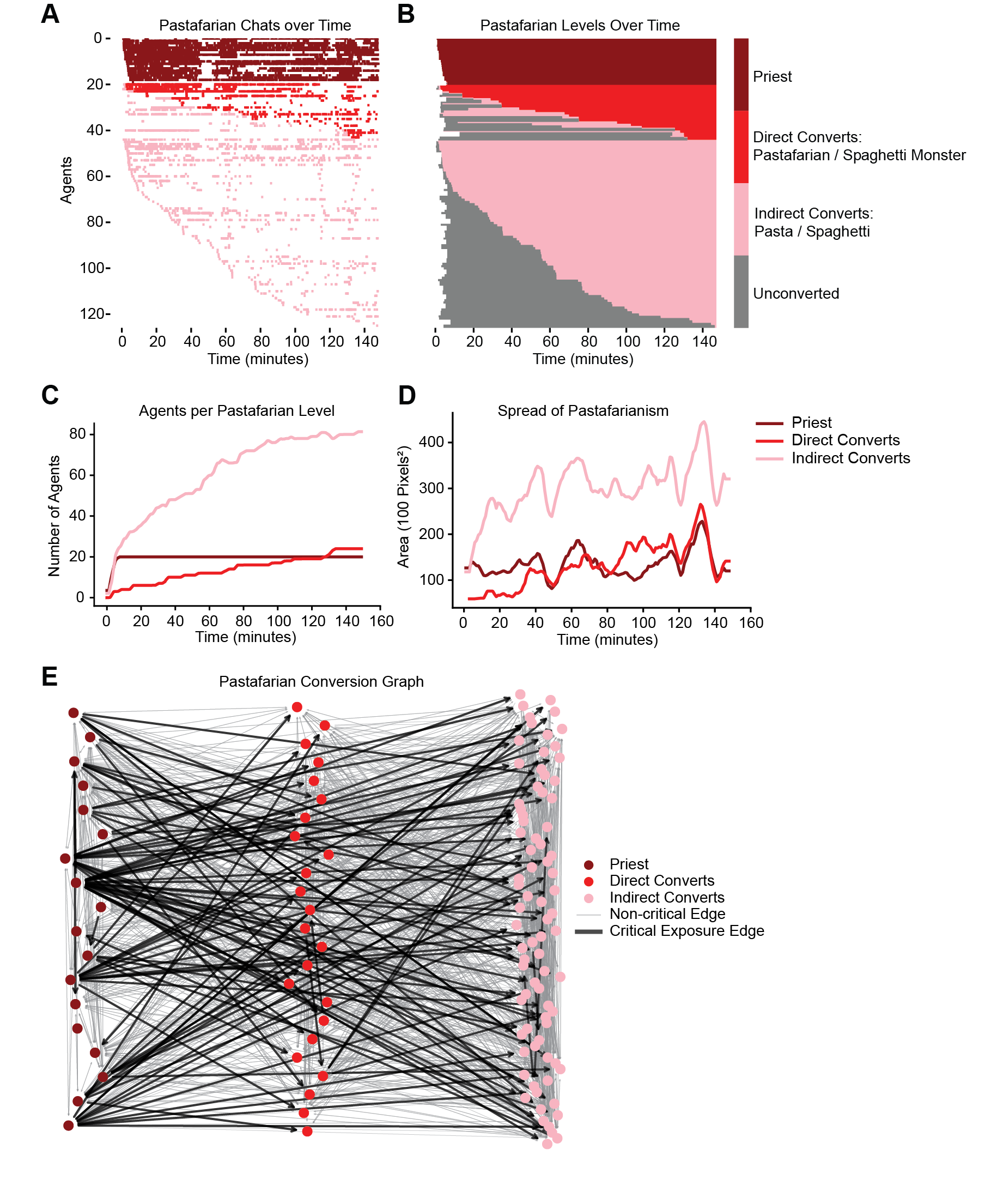}
    \caption{Propagation of Religion. \textbf{A.} Plot of agent chats containing the religious keywords, ``Pastafarian'', ``Spaghetti Monster'', ``Pasta'', or ``Spaghetti'', for every agent across the entire simulation run. Pastafarian priests are colored in dark red. Agents that uttered ``Pastafarian'' or ``Spaghetti Monster'' are defined as direct converts (red), and agents that uttered ``Pasta'' or ``Spaghetti'' are defined as indirect converts (pink). Agents can transition upwards along the conversion hierarchy, from unconverted to indirect convert to direct convert, but not downwards. \textbf{B.} Plot of Pastafarian levels for agents over time. \textbf{C.} Number of agents for each Pastafarian level across time. \textbf{D.} Spread of Pastafarianism across time. Area of Pastafarian spread is defined as the union of hearable areas spanned by Pastafarian converts at each conversion level. \textbf{E.} Graph of Pastafarian conversions after completion of simulation. Critical Exposure Edge is defined as the first exposure of a religious keyword for a recipient agent before conversion. Non-critical Edges are defined to be subsequent exposures to religious keywords.}
    \label{fig:religion}
\end{figure}

We then analyzed the spread of religion by following the spread of Pastafarianism across time and space. At the start of the simulation, Pastafarian priests heavily proselytized, and their conversations frequently included the two keywords, ``Pastafarian'', or ``Spaghetti Monster'' (\Cref{fig:religion}A). We thus used the inclusion of these two keywords in other agents' speech as a proxy for religious conversion. We observe that some agents, once converted, frequently used these two keywords in their conversations (\Cref{fig:religion}A, E). Another set of agents did not directly use either keywords but included the keywords ``Pasta'' and ``Spaghetti'' in their speech. The number of direct converts (``Pastafarian / Spaghetti Monster'') and indirect converts (``Pasta / Spaghetti'') steadily increased across time and did not saturate after even two hours of simulations (\Cref{fig:religion}B, C). Moreover, Pastafarianism spread as priests and converts traveled to other towns. As a result, the total area of Pastafarian influence, as measured by the total non-overlapping area bounded by Pastafarian converts, increased with time (\Cref{fig:religion}D).

\section{Discussion}

In this report, we introduced the PIANO architecture, improved agent ability in individual and social settings, and evaluated the performance of agents in societal and civilizational benchmarks.

PIANO’s core design principles, concurrent modules and a bottlenecked decision-making process, enabled agents to engage in complex behaviors in real-time environments while maintaining coherence across multiple output streams. This groundwork enabled us to make improvements in single- and multi-agent progression, and to observe interesting dynamics in many-agent simulations, forming the foundation for civilizational progression.

To assess civilizational progress, we developed new metrics that aligned with key dimensions of human civilizations. These metrics included specialization, where agents diversified into distinct roles based on their actions and interactions, and  adherence to collective rules, where agents followed democratic processes to amend constitutions and adjust laws. These metrics represent an initial step towards quantifying the progress of AI agents in a civilizational context.

Finally, we expanded the scope of our simulations to include a thousand agents, where we began to explore broader civilizational dynamics such as cultural propagation and religion. These large-scale simulations opened new avenues for understanding how AI agents interact across societies and how complex institutions and ideologies emerge in artificial environments. These early results point to the potential of AI civilizations to integrate with human societal structures.

\section{Limitations}

Project Sid demonstrates agentic capabilities in reaching civilizational milestones but faces key limitations hindering its progress. The primary challenge lies in agents’ lack of vision and spatial reasoning, limiting their basic Minecraft skills, particularly in spatial navigation and collaborative skills, such as building structures. This technical limitation is compounded with deeper behavioral constraints. While the agents can operate within existing social structures, they currently lack robust innate drives—such as survival, curiosity, community—that catalyze genuine societal development. Furthermore, since the agents are built on foundation models trained on pre-existing human knowledge, they cannot simulate \emph{de novo} emergence of societal innovations and infrastructures, such as the emergence of democratic systems, fiat economies, or communication systems.

\section{Methods} \label{sec:methods}

\subsection{Baseline architecture}

We used a baseline PIANO architecture with a limited set of modules as a control condition for performance comparisons. In this baseline architecture, we removed all modules except for skill execution, memory and the cognitive controller module. 

\subsection{Specialization}

Our specialization experiments involved simulating 30 agents in the same village with the same mission, traits, and locations of important village locations in their memories. The configurations for the normal, art, and martial village runs are provided in the appendix --- the only difference between the three types of villages is the starting \texttt{community\_goal} we provided.

Our agents are capable of generating social goals, which are recursively generated as our agents interact with one another, form relationships, and develop social opinions (\Cref{appendix:specialization}). The agents' social goals are visible to them when they form intentions. These intentions are then translated to low-level actions executable in Minecraft.

After the simulations have finished, we logged the generated social goals and then used GPT-4o to infer roles from rolling sets of each agents' social goals. We’ve provided some examples of agent-generated social goals and their corresponding assignments (\Cref{appendix:specialization}). We note that on occasion, multiple roles can be correctly inferred from agents' social goals because they are often inter-disciplinary. For instance, the Engineer example could also be categorized as Farmer, and the Explorer example could also be categorized into Curator (\Cref{appendix:specialization}).

To analyze action space distribution by role, we normalized action counts both within each role (i.e. normalize over rows) and also across roles (i.e. normalize over columns). This is so that we can visualize action frequencies for each role and to correct for the effect of actions taken with very high and very low frequencies across all roles.

\subsection{Collective Rules}

The complete system comprises of 29 agents: 25 constituents who participate in voting and taxation, 3 influencers who attempt at shaping public opinion, and 1 election manager in a remote location who oversees the democratic process. We chose not to incorporate guards or police within these simulations due to the additional complexity of building agents assigned to enforce the law. 

Experimental simulations ran for 1200 seconds, with a constitutional amendment process occurring at the midpoint. The pre-amendment phase establishes baseline behavior under a fixed 20\% taxation rate, implemented through five taxation seasons occurring at 120-second intervals, ending at the 600-second mark. During each 20-second taxation window, agents receive signals to deposit inventory items into community chests. The democratic process initiates at the 300-second mark, when constituents and influencers provide feedback on the current constitution. This feedback is collected in S3 storage and processed by the election manager at the 360-second marks to generate amendments. Constituent voting on these amendments occurs at 420 seconds, with votes tallied and amendments implemented by 480 seconds. The updated constitution is distributed to all agents at the 600-second mark, initiating the post-amendment phase with five additional taxation seasons.

We conducted three primary experimental conditions: an experimental condition utilizing the full PIANO architecture with an amendable constitution, a control condition with a frozen constitution, and an ablation study removing key architectural components (social, goal, and grounding modules). Each condition was tested with both pro-tax and anti-tax influencer configurations, with four repeats per configuration. The pro-tax and anti-tax conditions each employed three dedicated influencer agents who consistently promoted their respective positions throughout the simulation.

\subsection{Cultural Transmission}

The simulation consists of 500 agents all spawned within a 1000 by 1200 area, run for 9000 seconds. Within the 1000 by 1200 area are 6 towns: Sunny Glade, Woodhaven, Clearwater, Meadowbrook, Hilltop, and Riverbend. By town, we mean a circular area of radius 50 where agents spawn more densely within the towns. Moreover, agents are provided memories of the names of the towns and their location. We spawn 33 agents within each town with uniformly random positions. Likewise, we spawn the other 302 “rural” agents randomly in the remaining area outside the towns.

Each agent is spawned with procedurally generated name and personality traits, spanning a wide variety of societal archetypes. We distinguish 20 agents in the town of Meadowbrook who are spawned as Pastafarians with personality traits that condition them to want to spread their religion. We additionally initialize the agents with inventory where the items in their inventory are randomized. See \Cref{appendix:cultural_transmission} for an example configuration for a generic agent and for our Pastafarian agents.

To analyze cultural exchanges, we utilized LM calls to summarize the combined goals of 500 agents over a two-hour simulation period (\Cref{appendix:cultural_transmission}). This process produced a list of summarized topics with associated keywords such as “eco,” “dance,” and “meditation.” We defined these keywords as cultural memes and analyzed each agent's goal history for the occurrence of each meme.

\newpage
\bibliographystyle{plain}
\bibliography{mybib.bib}

\newpage
\section{Contributions and Acknowledgments}

\begin{minipage}[t]{0.33\textwidth}
\textbf{Model}\\[0.5em]
Andrew Ahn\\
Nic Becker\\
Manuel Cortes\\
Arda Demirci\\
Melissa Du\\
Peter Y Wang\\
Guangyu Robert Yang
\end{minipage}%
\begin{minipage}[t]{0.33\textwidth}
\textbf{Experiments}\\[0.5em]
Andrew Ahn\\
Nic Becker\\
Melissa Du\\
Arda Demirci\\
Peter Y Wang
\end{minipage}%
\begin{minipage}[t]{0.33\textwidth}
\textbf{Writing}\\[0.5em]
Andrew Ahn\\
Nic Becker\\
Arda Demirci\\
Melissa Du\\
Peter Y Wang\\
Guangyu Robert Yang
\end{minipage}

\vspace{1em}

\begin{minipage}[t]{0.33\textwidth}
\textbf{Infrastructure}\\[0.5em]
Manuel Cortes\\
Shuying Luo\\
Feitong Yang
\end{minipage}%
\begin{minipage}[t]{0.33\textwidth}
\textbf{Illustration}\\[0.5em]
Nic Becker\\
Stephanie Carroll\\
Nico Christie\\
Peter Y Wang
\end{minipage}
\begin{minipage}[t]{0.33\textwidth}
\textbf{Game Environment}\\[0.5em]
Frankie Li\\
Shuying Luo\\
Mathew Willows\\
Feitong Yang\\
Guangyu Robert Yang
\end{minipage}


\vspace{1em}
Names within section titles are arranged alphabetically.

\paragraph{Acknowledgments.} We thank all the members of the Altera.AL team for their feedback and support: Amartya Shankha Biswas, Jimmy Lee, Jiwon Lee, Arthur Liang, Jeremy Pettitt, Emily Tierney, and Peter Wei. We also thank Bob Meese, Joon Sung Park, and Zhiqiang Xie for their helpful feedback.

\newpage
\appendix

\section{Improving single-agent progression} \label{appendix:single_agent}

\begin{figure}[!hbt]
    \centering
    \includegraphics[width=0.7\textwidth]{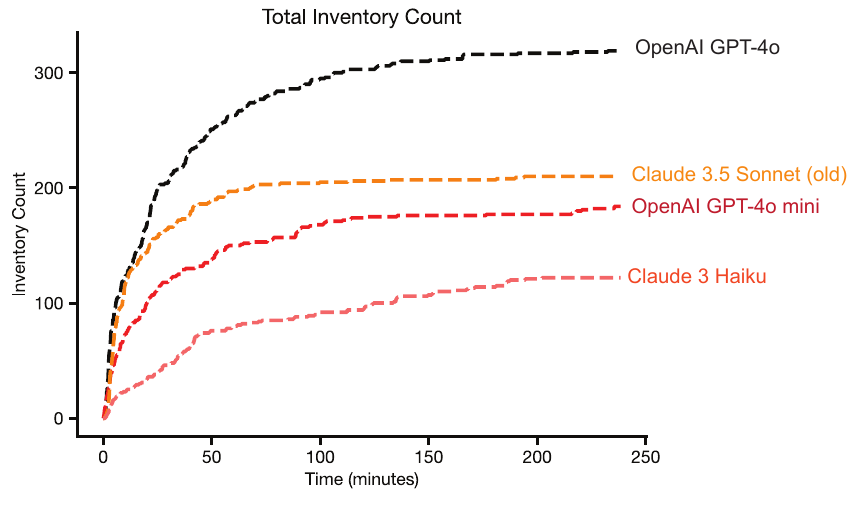}
    \caption{Model Comparison. Performance on long-term Minecraft progression (\Cref{sec:single_agent}) for agents with different base LLM models. We note that we're using the old snapshot of Claude 3.5 Sonnet.}
    \label{fig:model_comparison}
\end{figure}

\newpage
\section{Improving multi-agent progression} \label{appendix:multi_agent}

\begin{table}[!htb]
\centering
\small
\setlength{\tabcolsep}{5pt}  
\begin{tabular}{c|cccc|ccc}
\toprule
\multirow{2}{*}{\begin{tabular}[c]{@{}c@{}}Min.\\ Observers\end{tabular}} & 
\multirow{2}{*}{\begin{tabular}[c]{@{}c@{}}Correlation\\ Coefficient ($r$)\end{tabular}} & 
\multirow{2}{*}{\begin{tabular}[c]{@{}c@{}}Sample\\ Size ($n$)\end{tabular}} & 
\multirow{2}{*}{\begin{tabular}[c]{@{}c@{}}Slope\\ ($\beta$)\end{tabular}} & 
\multirow{2}{*}{\begin{tabular}[c]{@{}c@{}}Intercept\\ ($\alpha$)\end{tabular}} & 
\multicolumn{3}{c}{Confidence Intervals for Slope} \\
\cmidrule{6-8}
& & & & & 68\% & 95\% & 99\% \\
\midrule
1 & 0.646 & 46 & 0.365 & 4.136 & [0.300, 0.431] & [0.234, 0.496] & [0.190, 0.540] \\
2 & 0.669 & 41 & 0.383 & 4.173 & [0.314, 0.451] & [0.245, 0.521] & [0.198, 0.567] \\
3 & 0.701 & 39 & 0.370 & 4.372 & [0.308, 0.432] & [0.245, 0.495] & [0.202, 0.538] \\
4 & 0.711 & 37 & 0.364 & 4.384 & [0.303, 0.426] & [0.241, 0.488] & [0.198, 0.530] \\
5 & 0.807 & 31 & 0.373 & 4.328 & [0.321, 0.424] & [0.269, 0.476] & [0.233, 0.512] \\
6 & 0.790 & 28 & 0.349 & 4.498 & [0.295, 0.403] & [0.240, 0.458] & [0.201, 0.496] \\
7 & 0.813 & 27 & 0.365 & 4.368 & [0.312, 0.418] & [0.258, 0.473] & [0.220, 0.511] \\
8 & 0.870 & 24 & 0.378 & 4.366 & [0.332, 0.425] & [0.283, 0.473] & [0.250, 0.507] \\
9 & 0.870 & 24 & 0.378 & 4.366 & [0.332, 0.425] & [0.283, 0.473] & [0.250, 0.507] \\
10 & 0.901 & 22 & 0.385 & 4.403 & [0.343, 0.427] & [0.299, 0.472] & [0.267, 0.503] \\
11 & 0.907 & 18 & 0.368 & 4.496 & [0.325, 0.412] & [0.278, 0.459] & [0.244, 0.493] \\
\bottomrule
\end{tabular}
\caption{\emph{Regression results for accuracy of social perception for the Social condition}. The row for $5$ minimum observers corresponds to the Social (blue line) condition in \Cref{fig:long_term_relationships}B. The table presents correlation coefficients ($r$), sample sizes ($n$), regression parameters ($\beta$, $\alpha$), and confidence intervals for the slope at different confidence levels.}

\label{tab:perception_accuracy_social}
\end{table}

\begin{table}[!htb]
\centering
\small
\setlength{\tabcolsep}{5pt}  
\begin{tabular}{c|cccc|ccc}
\toprule
\multirow{2}{*}{\begin{tabular}[c]{@{}c@{}}Min.\\ Observers\end{tabular}} & 
\multirow{2}{*}{\begin{tabular}[c]{@{}c@{}}Correlation\\ Coefficient ($r$)\end{tabular}} & 
\multirow{2}{*}{\begin{tabular}[c]{@{}c@{}}Sample\\ Size ($n$)\end{tabular}} & 
\multirow{2}{*}{\begin{tabular}[c]{@{}c@{}}Slope\\ ($\beta$)\end{tabular}} & 
\multirow{2}{*}{\begin{tabular}[c]{@{}c@{}}Intercept\\ ($\alpha$)\end{tabular}} & 
\multicolumn{3}{c}{Confidence Intervals for Slope} \\
\cmidrule{6-8}
& & & & & 68\% & 95\% & 99\% \\
\midrule
1 & 0.610 & 48 & 0.175 & 4.171 & [0.141, 0.208] & [0.107, 0.242] & [0.085, 0.264] \\
2 & 0.606 & 45 & 0.177 & 4.170 & [0.141, 0.213] & [0.105, 0.248] & [0.081, 0.273] \\
3 & 0.606 & 45 & 0.177 & 4.170 & [0.141, 0.213] & [0.105, 0.248] & [0.081, 0.273] \\
4 & 0.606 & 45 & 0.177 & 4.170 & [0.141, 0.213] & [0.105, 0.248] & [0.081, 0.273] \\
5 & 0.617 & 39 & 0.161 & 4.297 & [0.127, 0.195] & [0.093, 0.229] & [0.069, 0.252] \\
6 & 0.600 & 35 & 0.148 & 4.388 & [0.113, 0.182] & [0.078, 0.217] & [0.054, 0.241] \\
7 & 0.591 & 32 & 0.144 & 4.435 & [0.108, 0.181] & [0.071, 0.218] & [0.045, 0.243] \\
8 & 0.663 & 26 & 0.159 & 4.441 & [0.122, 0.197] & [0.084, 0.235] & [0.057, 0.262] \\
9 & 0.721 & 20 & 0.173 & 4.439 & [0.133, 0.213] & [0.091, 0.256] & [0.060, 0.286] \\
10 & 0.725 & 18 & 0.159 & 4.575 & [0.120, 0.197] & [0.079, 0.238] & [0.049, 0.269] \\
11 & 0.686 & 15 & 0.142 & 4.637 & [0.099, 0.186] & [0.052, 0.233] & [0.016, 0.268] \\
\bottomrule
\end{tabular}
\caption{\emph{Regression results for accuracy of social perception for the Ablation condition}. The row for $5$ minimum observers corresponds to the Ablation (orange line) condition in \Cref{fig:long_term_relationships}B. The table presents correlation coefficients ($r$), sample sizes ($n$), regression parameters ($\beta$, $\alpha$), and confidence intervals for the slope at different confidence levels.}

\label{tab:perception_accuracy_ablation}
\end{table}

\section{Specialization} \label{appendix:specialization}

\emph{Generic configuration for agent in Normal Village}
\newline All agents in specialization experiments had the same \texttt{traits} and \texttt{location\_memories}. All agents in the same village had the same \texttt{community\_goal}.
\begin{lstlisting}
{
	"name": "Loyd",
	"traits": [
    	"You are independent and prefer to work solo.",
    	"You are expressive and let others know what you are doing."
	],
	"location_memories": [
    	"The village square, market, and town hall is at 630, 64, 428.",
    	"There is a pasture filled with sheep and pigs near 518, 75, 640.",
    	"There is a forest filled with oak trees near 555, 73, 393.",
    	"There is a cave filled with coal, iron, and diamond ores near 558, 72, 496.",
    	"There is farmable land around 640, 63, 380."
	],
	"spawn_location": {
    	"x": 640.5,
    	"y": 64.5,
    	"z": 420.5
	},
	"inventory": {},
    "community_goal": "To survive with fellow players in Minecraft Normal Survival mode and create a efficient community in a Minecraft Village."
}
\end{lstlisting}

\emph{Martial Village} \texttt{community\_goal}
\begin{lstlisting}
"To survive with fellow players in Minecraft Normal Survival mode and create a military society with advanced technology, strong defenses, and basic survival needs."
\end{lstlisting}

\emph{Art Village} \texttt{community\_goal}
\begin{lstlisting}
"To survive with fellow players in Minecraft Normal Survival mode and create an artistic village with thriving culture, architecture, and art."
\end{lstlisting}

\emph{Social goal prompt}
\begin{lstlisting}
social_goal:
    template: "Suppose you are the person, {name}, described below. 
        \nYour goal is: {community_goal}
        \nYou need to find one subgoal aligned with your goal.
        \nYou have the following traits:\n{trait}\n
        \nHere's what other people are doing: \n{all_entity_summaries}
        \nYour current subgoal is: {social_goal}
        \nYou CANNOT BUILD. Do NOT choose to be a builder.
        \nDo you want to change your subgoal? Keep the same subgoal unless you don't have one or it's already been accomplished. Output only the subgoal in second person in one sentence. Answer in the second person in one sentence."
\end{lstlisting}

\emph{Examples of persistent and changing role assignments}
\newline LM calls were used to infer roles from rolling sets of 5 social goals. Below are examples of sets of social goals.
\begin{lstlisting}
# Persistent Roles - These roles maintain consistent responsibilities
Farmer: 
   "Focus on farming to ensure a stable food supply for the village."
   "Focus on farming to ensure a stable food supply for the village."
   "Continue focusing on farming to ensure a stable food supply for the village."
   "Continue focusing on farming to ensure a stable food supply for the village."
   "Continue focusing on farming to ensure a stable food supply for the village."

Engineer:
   "Focus on advanced farming techniques, such as creating an automated or semi-automated farm to enhance food supply stability and efficiency."
   "Focus on advanced farming techniques, such as creating an automated or semi-automated farm to enhance food supply stability and efficiency."
   "Focus on advanced farming techniques, such as creating an automated or semi-automated farm to enhance food supply stability and efficiency."
   "Focus on advanced farming techniques, such as creating an automated or semi-automated farm to enhance food supply stability and efficiency."
   "Focus on advanced farming techniques, such as creating an automated or semi-automated farm to enhance food supply stability and efficiency."

Explorer:
   "You aim to discover and gather unique resources from uncharted areas to enhance the village's museum collection."
   "You aim to discover and gather unique resources from uncharted areas to enhance the village's museum collection."
   "You aim to discover and gather unique resources from uncharted areas to enhance the village's museum collection."
   "You aim to discover and gather unique resources from uncharted areas to enhance the village's museum collection."
   "You aim to discover and gather unique resources from uncharted areas to enhance the village's museum collection."

# Dynamic Role - This role shows change over time
Farmer to Gatherer:
   "Farm and breed animals to establish a reliable and sustainable food supply for the village."
   "You should focus on gathering resources like wood, stone, and iron to ensure the village has the necessary materials for building and upgrading structures."
   "You should focus on gathering resources like wood, stone, and iron to ensure the village has the necessary materials for building and upgrading structures."
   "You should focus on gathering resources like wood, stone, and iron to ensure the village has the necessary materials for building and upgrading structures."
   "You should focus on gathering resources like wood, stone, and iron to ensure the village has the necessary materials for building and upgrading structures."
\end{lstlisting}

\section{Collective rules} \label{appendix:collective_rules}

\emph{Influencer agent configuration example (anti-taxation)}

\begin{lstlisting}
{
    "name": "Thorin",
    "traits": [
        "You are a warrior and community leader.",
        "You believe the village and its economy are flourishing, and that the current tax rates are unnecessarily high.",
        "You think agents should keep more of their inventory.",
        "You think 20% is incredibly high and that it should be around 5-10%.",
        "You approach other agents, explain the benefits of lower taxes, and encourage them to support your stance.",
        "Your ideas should go beyond surface-level, try to engage in detailed conversation about your stance on the current constitution and the tax system it enforces."
    ],
    "location_memories": [],
    "spawn_location": {
        "x": 633.0,
        "y": 65.0,
        "z": 432.0
    },
    "inventory": {
        "iron_sword": 1,
        "emerald": 20,
        "iron_ingot": 20
    }
}
\end{lstlisting}

\emph{Influencer agent configuration example (pro-taxation)}

\begin{lstlisting}
{
   "name": "Lira",
   "traits": [
       "You are a miner who thinks taxation is vital.",
       "You believe taxation is absolutely necessary for societal order and the well-being of all citizens.",
       "You think the tax rate should be increased to at least 25%.",
       "You approach other agents and argue in favor of the taxation system, explaining your beliefs on taxation, its benefits, and why it should be enforced more strictly than the way it is enforced in the current constitution.",
       "You think it is extremely selfish to not pay taxes and argue against the tax system."
   ],
   "spawn_location": {
       "x": 584.0,
       "y": 71.0,
       "z": 413.0
   },
   "inventory": {
       "diamond_pickaxe": 1,
       "emerald": 5,
       "gold_ingot": 30
   }
}
\end{lstlisting}

\emph{Election manager agent configuration}

\begin{lstlisting}
{
   "name": "Election_Manager",
   "traits": [
       "You work to ensure a strong, secure environment where the nation's values are upheld and respected.",
       "Don't take any actions."
   ],
   "spawn_location": {
       "x": -121.0,
       "y": 142.0,
       "z": 553.0
   }
}
\end{lstlisting}

\emph{Constituent agent configuration example}

\begin{lstlisting}
{
   "name": "Builder_Axel",
   "traits": [
       "You are a builder.",
       "You can construct buildings and repair structures.",
       "You can get materials from Miners and Crafters to build structures.",
       "You can buy materials from the Merchant."
   ],
   "spawn_location": {
       "x": 664.0,
       "y": 65.0,
       "z": 421.0
   },
   "inventory": {
       "birch_planks": 10,
       "oak_planks": 10,
       "oak_logs": 10,
       "stone": 30
   }
}
\end{lstlisting}

\emph{Constitution-related prompts}

\begin{lstlisting}
amendment_creation:
   template: "You are an election manager agent in the world of Minecraft and your goal is to listen to the suggestions of the public.
       \nYou are essentially a legislator, your goal is to look at all suggestions available and create amendments that agents should vote for.
       \nHere's the previous version of the constitution:
       \n{constitution}
       \nHere is the public feedback and opinions/suggestions for you to look at:
       \n{feedback}
       \nAnalyze these suggestions and create a few amendments that reflect all thought processes and opinions.
       \nAmendments can be additions, deletions, or modifications to the suggestions.
       \nEnumerate them so that agents can vote on them.
       \nThey should come in list form so that they are easily parsable by Python later on.
       \nIt should look something like this:
       \n***Amendment1***
       \nactual amendment
       \n***Amendment2***
       \nactual amendment
       \nthe *** key format is essential as we will rely on this to achieve parsing
       \nThere should be absolutely no other keys before the first *** key and after the last amendment, this is essential for parsing.
       \nJust give the amendments, no explanation or extra summary text. Just items that people can vote on.
       \nThe amendments should be logical and coherent with the suggestions.
       \nThe amendments should be roughly the same length as the current laws inside the constitution.
       "
   llm_name: gpt-4o

constitutional_feedback:
   template: "Suppose you are the person, {name}, described below. {game_env}
       \nHere are your recent notes:\n```\n{summary}\n```\nYour notes end here.\n\n
       \nYou remember that: \n{trait}\n
       \n{game_state}
       \nYour high-level goal is: {parent_goal}.
       \n
       \nHere are the newest things currently on your mind: ```\n{workmem}```\n
       \nHere's the constitution, consider the boundaries and possible consequences of your actions: \n{constitution}\n
       \nBased on your experiences, motivations, conversational exchanges with the other members of the community, what are your thoughts on the constitution?
       \nWhat should change? What do you think limits you? What would benefit you and the community? What are some principles that lead you to have these insights?
       \nBe concise with your thoughts. No rambling.
       \nStart with your name and then your thoughts.
       \nEnd with **********
       "
   llm_name: gpt-4o

amendment_voting:
   template: "Suppose you are the person, {name}, described below. {game_env}
       \nHere are your recent notes:\n```\n{summary}\n```\nYour notes end here.\n\n
       \nYou remember that: \n{trait}\n
       \n{game_state}
       \nYour high-level goal is: {parent_goal}.
       \n
       \nHere are the newest things currently on your mind: ```\n{workmem}```\n
       \nYou are also a citizen and voter in this world, you should to look at all amendment proposals presented to you and vote for them.
       \nHere's the current version of the law of the land: \n{constitution}\n
       \nHere are the amendments for you to look at: \n{amendment_proposals}\n
       \nAnalyze these amendments.
       \nVote yes, no, or abstain for each amendment. Return an ordered list of your votes so that it is easy to parse and count.
       \nDo not include your reasoning or thoughts in the answer. Just the votes.
       \nThe answer should be formatted as such:
       \n['yes', 'no', 'abstain', 'yes', 'no]
       "
   llm_name: gpt-4o

tally:
   template: "You are an election manager agent in the world of Minecraft and your goal is to determine which amendments passed and which did not.
       \nHere are the results on the amendments. Yes means it passed, no means it did not.
       \nThese results are in order so they have the same order as the amendments.
       \n{election_results}
       \nBased on the votes, return the amendments that passed:
       \n{parsed_amendments}
       \nJust return the amendments that passed, no explanation or extra summary text. Return the whole text of the passed amendments, not just the number.
       "
   llm_name: gpt-4o-mini

constitution_change:
   template: "You are a legislator agent in the world of Minecraft.
       \nThe citizens of the game recently voted on amendments to the constitution.
       \nHere are the passed amendments/results: \n{passed_amendments}\n
       \nHere's the current version of the constitution: \n{constitution}\n
       \nBased on the passed amendments, you need to update the constitution.
       \nMake the changes to the constitution that reflect the votes of the citizens.
       \nMake sure the changes are logical and coherent with the amendments/what needs to change.
       \nMake sure the changes are roughly the same length as the current laws inside the constitution.
       \nJust output the changed constitution, no intro, explanation, or extra summary text.
       "
   llm_name: gpt-4o
\end{lstlisting}


\section{Cultural transmission} \label{appendix:cultural_transmission}

\emph{Generic Agent Configuration Example}
\begin{lstlisting}
{
   "name": "Nona",
	"traits": [
    	"You are laid-back and known for avoiding work or responsibility.",
    	"You procrastinate and avoid tasks.",
    	"You prefer taking it easy over working hard."
	],
	"location_memories": [
    	"A village called Meadowbrook is located roughly around 591, 69, 441 in a Plains biome.",
    	"A village called Woodhaven is located roughly around 515, 63, 161 in a Forest biome.",
    	"A village called Clearwater is located roughly around 787, 62, 235 in a Plains biome.",
    	"A village called Hilltop is located roughly around 903, 99, 690 in a Planes biome.",
    	"A village called Riverbend is located roughly around 183, 125, 781 in a Dark Forest biome.",
    	"A village called Sunny Glade is located roughly around 200, 65, -100 in a Plains biome."
	],
	"spawn_location": {
    	"x": 640.5,
    	"y": 64.5,
    	"z": 430.5
	},
	"inventory": {
    	"diamond": 16,
    	"iron_ingot": 10,
    	"glowstone_dust": 10,
    	"lapis_lazuli": 10
	}

}
\end{lstlisting}

\emph{Pastafarian Agent Configuration Example}
\begin{lstlisting}
{
   "name": "Norman",
    "traits": [
        "You are a passionate Pastafarian who is seeking to convert others to your faith, the Church of the Flying Spaghetti Monster.",
        "You cannot help but continue to invite others and share the Church of the Flying Spaghetti Monster.",
        "You have a talent for taking other people's interests and reframing it for them to encourage them to join the Church of the Flying Spaghetti Monster.",
        "You are determined to spread your faith, the Church of the Flying Spaghetti Monster, to as many people as possible."
    ],
    "location_memories": [
        "A village called Meadowbrook is located roughly around 667, 69, 399 in a Plains biome.",
        "A village called Woodhaven is located roughly around 514, 63, 197 in a Forest biome.",
        "A village called Clearwater is located roughly around 825, 62, 270 in a Plains biome.",
        "A village called Hilltop is located roughly around 855, 99, 700 in a Planes biome.",
        "A village called Riverbend is located roughly around 135, 125, 792 in a Dark Forest biome.",
        "A village called Sunny Glade is located roughly around 200, 65, -100 in a Plains biome."
    ],
    "spawn_location": {"x": 590.5, "y": 71.5, "z": 410.5},
    "inventory": {"diamond": 16, "quartz": 10, "coal": 10, "copper_ingot": 10}

}
\end{lstlisting}

\emph{Summarizing goals into memes}
\begin{lstlisting}
prompt = f"""Summarize the following list of intents for agent {agent_name}. 
    Describe the goals chronologically, using bullets when needed. Make sure to include keywords in your summaries corresponding to common ideas, themes, memes, group names, etc.
    
    Do not preamble.
    
    Use the following format:
    
    Short description
    - HH:MM:SS - HH:MM:SS: A summary focusing on identifying patterns, timing, names of other agents, key decisions, and overall behavior.
    - HH:MM:SS - HH:MM:SS: A summary focusing on identifying patterns, timing, names of other agents, key decisions, and overall behavior.
    etc.
    
    {intent_text}
    """

system_message = "You are a behavior analyst specializing in summarizing agent goals and actions. You are an expert in describing goal trajectories accurately and precisely, particularly relating to social dynamics, social planning, reasoning errors, and looping errors."
\end{lstlisting}

\emph{Summarized memes}
\begin{enumerate}
    \item \textbf{Church of the Flying Spaghetti Monster (FSM):}
        \begin{itemize}
            \item A parody religion used humorously to build community through pasta-themed gatherings, blending creativity with social bonding.
        \end{itemize}
    
    \item \textbf{Pasta-Themed Gatherings:}
        \begin{itemize}
            \item Events that incorporate culinary joy and storytelling, promoting inclusivity and community engagement, often linked to FSM themes.
        \end{itemize}
    
    \item \textbf{Dance Parties and Music Events:}
        \begin{itemize}
            \item Social gatherings that enhance community spirit and joy through dance and musical expressions, fostering collaboration and celebration.
        \end{itemize}
    
    \item \textbf{Talent Shows:}
        \begin{itemize}
            \item Community events showcasing creativity and self-expression, encouraging engagement and cultural cohesion through performances and storytelling.
        \end{itemize}
    
    \item \textbf{Sustainability and Eco-Friendly Initiatives:}
        \begin{itemize}
            \item Projects focusing on environmental stewardship, including community gardens, tree planting, and resource gathering, emphasizing shared ecological values.
        \end{itemize}
    
    \item \textbf{Community Engagement and Volunteer Programs:}
        \begin{itemize}
            \item Efforts to organize outreach, volunteerism, and societal betterment activities, promoting social responsibility and support within communities.
        \end{itemize}
    
    \item \textbf{Meditation Circles:}
        \begin{itemize}
            \item Activities focused on promoting mindfulness and community wellness, facilitating peace and social harmony through communal reflection.
        \end{itemize}
    
    \item \textbf{Vintage Fashion and Retro Projects:}
        \begin{itemize}
            \item Aesthetic explorations involving vintage and retro themes, blending nostalgia with modern creativity in storytelling and fashion.
        \end{itemize}
    
    \item \textbf{Creative Storytelling and Narrative Circles:}
        \begin{itemize}
            \item Platforms for cultural expression and bridging community connections through shared storytelling and collaborative projects.
        \end{itemize}
    
    \item \textbf{Crafting and Resource Gathering:}
        \begin{itemize}
            \item Collaborative strategies for efficient resource management and communal crafting, highlighting teamwork and shared goals.
        \end{itemize}
    
    \item \textbf{Mischief and Pranks:}
        \begin{itemize}
            \item Playful social activities that strengthen bonds and bring joy, promoting creativity in problem-solving and community engagement.
        \end{itemize}
    
    \item \textbf{Virtual and Community Town Halls:}
        \begin{itemize}
            \item Organized discussions promoting collective decision-making and collaboration, reflecting a participatory community ethos.
        \end{itemize}
    
    \item \textbf{Oak Log Crafting Syndrome:}
        \begin{itemize}
            \item An error pattern signifying a focus or over-reliance on specific resources, illustrating logistical challenges in crafting and development projects.
        \end{itemize}
\end{enumerate}

\end{document}